\documentclass[11pt,a4paper]{article}
\usepackage[hyperref]{acm2019}
\usepackage{times}
\usepackage{latexsym}
\usepackage{breakurl}
\hypersetup{breaklinks=true}

\aclfinalcopy 
\def\aclpaperid{450} 

\usepackage{amsmath,amsfonts,amsthm}
\usepackage{pbox}
\usepackage{color}
\usepackage{graphicx}
\usepackage{pifont}
\newcommand{\cmark}{\ding{51}}%
\newcommand{\xmark}{\ding{55}}%
\usepackage{footnote}
\makesavenoteenv{tabular}
\makesavenoteenv{table}

\newtheoremstyle{mydef}%
{}{}{\normalfont}{}{\itshape}{.\,}{ }{}
\theoremstyle{mydef}

\newtheoremstyle{myprob}%
{}{}{\normalfont}{}{\scshape}{.\,}{ }{}
\theoremstyle{myprob}
\newtheorem{problem}{Problem}

\newtheoremstyle{mylayer}{}{}{\normalfont}{}{\itshape}{.\,}{ }{}\theoremstyle{mylayer}
\newtheorem{layer1}{}

\newtheoremstyle{mylayer}{}{}{\normalfont}{}{\itshape}{.\,}{ }{}\theoremstyle{mylayer}

\begin{document}

\newcommand\BibTeX{B{\sc ib}\TeX}

\title{Multi-Style Generative Reading Comprehension}

\author{Kyosuke Nishida$^1$, 
Itsumi Saito$^1$, 
Kosuke Nishida$^1$, 
\\{\bf Kazutoshi Shinoda}$^2$\thanks{\ \ Work done during an internship at NTT.}, 
{\bf Atsushi Otsuka}$^1$,
{\bf Hisako Asano}$^1$, 
{\bf Junji Tomita}$^1$\\
  $^1$NTT Media Intelligence Laboratory, NTT Corporation \hspace{1.5em}  $^2$The University of Tokyo\\
  {\tt kyosuke.nishida@acm.org}
}

\date{}

\maketitle

\begin{abstract}
This study tackles generative reading comprehension (RC), which consists of answering questions based on textual evidence and natural language generation (NLG).
We propose a multi-style abstractive summarization model for question answering, called Masque.
The proposed model has two key characteristics.
First, unlike most studies on RC that have focused on extracting an answer span from the provided passages, our model instead focuses on generating a summary from the question and multiple passages.
This serves to cover various answer styles required for real-world applications. 
Second, 
whereas previous studies built 
a specific model for each answer style because of the difficulty of acquiring one general model,
our approach learns multi-style answers within a model to improve the NLG capability for all styles involved.
This also enables our model to 
give an answer in the target style.
Experiments show that our model achieves state-of-the-art performance on 
the Q\&A task and the Q\&A + NLG task 
of MS MARCO 2.1 and the summary task of NarrativeQA. 
We observe that the transfer of the style-independent NLG capability to the target style is the key to its success.
\end{abstract}

\section{Introduction}
\label{sec:intro}

Question answering has been a long-standing research problem. Recently, reading comprehension (RC), a challenge to answer a question given textual evidence provided in a document set, has received much attention. 
Current mainstream studies have treated RC as a process of extracting an answer span from one passage~\citep{RajpurkarZLL16,RajpurkarJL18} or multiple passages~\citep{JoshiCWZ17,Yang0ZBCSM18}, which is usually done by predicting the start and end positions of the answer~\citep{Yu18,DevlinCLT18}. 

The demand for answering questions in natural language is increasing rapidly, and this has led to the development of smart devices such as Alexa. In comparison with answer span extraction, however, the natural language generation (NLG) capability for RC has been less studied. While datasets such as MS MARCO~\citep{Bajaj18} and NarrativeQA~\citep{KociskySBDHMG18} have been proposed for providing abstractive answers, the state-of-the-art methods for these datasets
are based on answer span extraction~\citep{WuWLHWLLL18,HuPWHLYZ18}.  
Generative models 
suffer from a dearth of training data to cover open-domain questions.


\begin{figure}[t!]
\centering
\includegraphics[width=.48\textwidth]{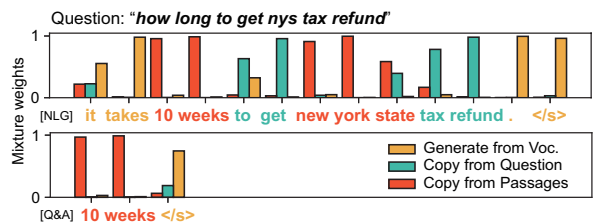}
\caption{Visualization of how our model generates an answer on MS MARCO.
Given an answer style (\textbf{top}:~NLG, \textbf{bottom}:~Q\&A), the model controls the mixture of three distributions for generating words from a vocabulary and copying words from the question and multiple passages at each decoding step.
}
\label{fig:mixratio}
\end{figure}
 
Moreover, to satisfy various information needs, intelligent agents should be capable of answering \textit{one} question in \textit{multiple} styles, such as well-formed sentences,
which make sense even without the context of the question and passages,
and concise phrases. These capabilities complement each other, but 
previous studies
cannot use and control different styles within a model.

In this study, we propose \textit{Masque}, a generative model for multi-passage RC. It achieves state-of-the-art performance
on the Q\&A task and the Q\&A + NLG task 
of MS MARCO 2.1 and the summary task of NarrativeQA. 
The main contributions of this study are as follows.

\paragraph{Multi-source abstractive summarization.} 

We introduce the pointer-generator mechanism~\citep{SeeLM17}
for generating an abstractive answer from the question and multiple passages,
which covers
various answer styles.
We extend the mechanism to a Transformer~\citep{VaswaniSPUJGKP17} based one that allows words to be generated from a vocabulary and to be copied from the question and passages.

\paragraph{Multi-style learning for style control and transfer.} 
We introduce multi-style learning that enables our model to control answer styles
and improves RC
for all styles involved. 
We also extend the pointer-generator 
to a conditional decoder by introducing an artificial token corresponding to each style, as in~\citep{JohnsonSLKWCTVW17}.  For each decoding step, it controls the mixture weights over three  distributions with the given style (Figure~\ref{fig:mixratio}).

\section{Problem Formulation}
\label{sec:problem}

This paper considers the following task:
\begin{problem}
\label{prob:prob}
Given a question with $J$ words $x^q = \{x^q_1, \ldots, x^q_J\}$, a set of $K$ passages, where the $k$-th passage is composed of $L$ words $x^{p_k} = \{x^{p_k}_1, \ldots, x^{p_k}_{L}\}$, and an answer style label $s$, an RC model outputs an answer $y = \{y_1, \ldots, y_T \}$ conditioned on the style.
\end{problem}
In short, 
given a 3-tuple $(x^q, \{x^{p_k}\}, s)$, the system predicts $P(y)$.
The training data is a set of 6-tuples: $(x^q, \{x^{p_k}\}, s, y, a, \{r^{p_k}\})$, where $a$ and $\{r^{p_k}\}$ are optional.
Here, $a$ is $1$ if the question is answerable with the provided passages and $0$ otherwise, and 
$r^{p_k}$ is $1$ if the $k$-th passage is required to formulate the answer and $0$ otherwise.

\section{Proposed Model} 

\begin{figure}[t!]
\centering
\includegraphics[width=.47\textwidth]{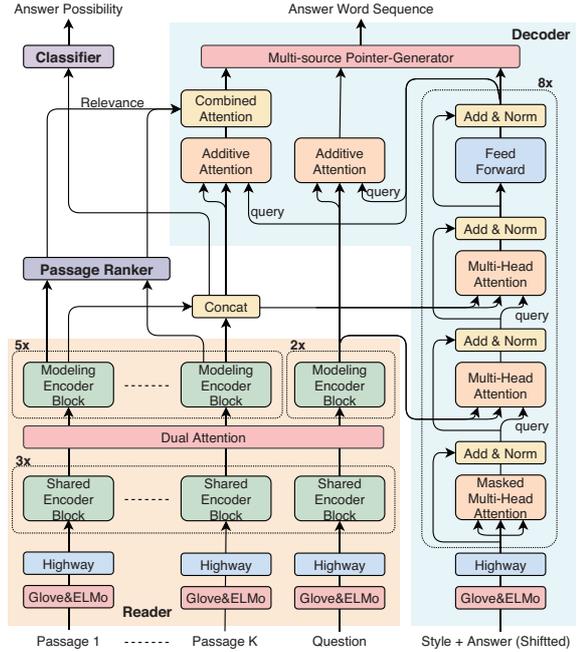}
\caption{Masque model architecture.}
\label{fig:model}
\end{figure}

We propose a Multi-style Abstractive Summarization model for QUEstion answering, called \textit{Masque}.
Masque directly models the conditional probability $p(y|x^q, \{x^{p_k}\}, s)$.
As shown in Figure~\ref{fig:model}, it consists of the following modules.
\begin{layer1}
The \textbf{question-passages reader} (\S\ref{sec:reader}) models interactions between the question and passages.
\end{layer1}
\begin{layer1}
The \textbf{passage ranker} (\S\ref{sec:ranker}) finds passages relevant to the question.
\end{layer1}
\begin{layer1}
The \textbf{answer possibility classifier} (\S\ref{sec:classifier}) identifies answerable questions.
\end{layer1}
\begin{layer1}
The \textbf{answer sentence decoder} (\S\ref{sec:decoder}) outputs an answer sentence conditioned on the target style.
\end{layer1}

Our model is based on multi-source abstractive summarization: the answer that it generates can be viewed as a summary from the question and passages. 
The model also learns multi-style answers together.
With these two characteristics, we aim to acquire the style-independent NLG ability and transfer it to the target style.
In addition, to improve natural language understanding in the reader module, our model considers RC, passage ranking, and answer possibility classification together as multi-task learning.

\subsection{Question-Passages Reader}
\label{sec:reader}

The reader module is shared among multiple answer styles and the three task-specific modules.

\subsubsection{Word Embedding Layer}

Let $x^q$ and $x^{p_k}$ represent 
one-hot vectors (of size $V$) for 
words in the question and the $k$-th passage. First, this layer projects each of the vectors to a $d_\mathrm{word}$-dimensional vector with a pre-trained weight matrix $W^e \in \mathbb{R}^{d_\mathrm{word} \times V}$ such as GloVe~\citep{PenningtonSM14}. Next, it uses contextualized word representations via ELMo~\citep{PetersNIGCLZ18}, which 
allows our model to use morphological clues to form robust representations for out-of-vocabulary words unseen in training. 
Then, the concatenation of the word and contextualized vectors is passed to a two-layer highway network~\citep{SrivastavaGS15} to fuse the two types of embeddings, as in \citep{SeoKFH17}.
The highway network is shared 
by the question and passages. 

\subsubsection{Shared Encoder Layer}
\label{sec:tfenc}

This layer uses a stack of Transformer blocks, which are shared by the question and passages, on top of the embeddings provided by the word embedding layer. The input of the first block is immediately 
mapped to a $d$-dimensional vector by a linear transformation.
The outputs of this layer are $E^{p_k} \in \mathbb{R}^{d \times L}$ for each $k$-th passage, and $E^q \in \mathbb{R}^{d \times J}$ for the question.

\paragraph{Transformer encoder block.} The block consists of two sub-layers: a self-attention layer and a position-wise feed-forward network. For the self-attention layer, we adopt the multi-head attention mechanism~\citep{VaswaniSPUJGKP17}. 
Following GPT~\citep{RadfordNSS18}, the feed-forward network consists of two linear transformations with a GELU~\citep{HendrycksG16} activation function in between.
Each sub-layer is placed inside a residual block~\citep{HeZRS16}. For an input $x$ and a given sub-layer function $f$, the output is $\mathrm{LN}(f(x)+x)$, where $\mathrm{LN}$ indicates the layer normalization~\citep{BaKH16}.
To facilitate these residual connections, 
all sub-layers produce a sequence of $d$-dimensional vectors.
Note that our model does not use any position embeddings in this block because ELMo gives the positional information of the words in each sequence.

\subsubsection{Dual Attention Layer}
\label{sec:dual}

This layer uses a dual attention mechanism to fuse information from the question to the passages 
as well as from the passages to the question. 

It first computes a similarity matrix $U^{p_k} \in \mathbb{R}^{L{\times}J}$ between the question and the $k$-th passage, as done in \citep{SeoKFH17}, where
\begin{align}
\nonumber
U^{p_k}_{lj} = {w^a}^\top [ E^{p_k}_l; E^q_j; E^{p_k}_l \odot E^q_j ]
\end{align}
indicates the similarity between the $l$-th word of the $k$-th passage and the $j$-th question word. The $w^a \in \mathbb{R}^{3d}$ are learnable parameters. The $\odot$ operator denotes the Hadamard product, and the $[;]$ operator denotes vector concatenation across the rows. Next, the layer obtains 
the row and column 
normalized similarity matrices $A^{p_k} = \mathrm{softmax}_j({U^{p_k}}^\top)$
and $B^{p_k}  = \mathrm{softmax}_{l}(U^{p_k})$.
It then uses DCN~\citep{XiongZS17} 
to obtain dual attention representations, $G^{q \rightarrow p_k} \in \mathbb{R}^{5d \times L}$ and $G^{p \rightarrow q} \in \mathbb{R}^{5d \times J}$:
\begin{align}
\nonumber
G^{q \rightarrow p_k} &= [E^{p_k}; \bar{A}^{p_k}; \bar{\bar{A}}^{p_k}; E^{p_k} \odot \bar{A}^{p_k}; E^{p_k} \odot \bar{\bar{A}}^{p_k}] \\
\nonumber
G^{p \rightarrow q} &= [ E^{q} ; \bar{B}; \bar{\bar{B}}; 
 E^{q} \odot \bar{B}; E^{q} \odot \bar{\bar{B}} ].
\end{align}
Here, $\bar{A}^{p_k} =  E^q A^{p_k}$, 
$\bar{B}^{p_k} =  E^{p_k} B^{p_k}$, 
$\bar{\bar{A}}^{p_k} = \bar{B}^{p_k} A^{p_k}$, 
$\bar{\bar{B}}^{p_k} = \bar{A}^{p_k} B^{p_k}$, 
$\bar{B} = \max_k(\bar{B}^{p_k})$, and
$\bar{\bar{B}} = \max_k(\bar{\bar{B}}^{p_k})$.

\subsubsection{Modeling Encoder Layer}

This layer uses a stack of the Transformer encoder blocks for question representations and obtains $M^q \in \mathbb{R}^{d \times J}$ from $G^{p \rightarrow q}$. It also uses another stack for passage representations and obtains $M^{p_k} \in \mathbb{R}^{d \times L}$ from $G^{q \rightarrow p_k}$ for each $k$-th passage. The outputs of this layer, $M^q$ and $\{M^{p_k}\}$, are passed on to the answer sentence decoder; the $\{M^{p_k}\}$ are also passed on to the passage ranker and the answer possibility classifier.

\subsection{Passage Ranker}
\label{sec:ranker}

The ranker maps the output of the modeling layer,  $\{M^{p_k}\}$, to the relevance score of each passage. 
It takes the output for the first word, $M^{p_k}_1$, which corresponds to the beginning-of-sentence token, to obtain the aggregate representation of each passage sequence.
Given $w^r \in \mathbb{R}^{d}$ as learnable parameters, it calculates the relevance of each $k$-th passage to the question as
\begin{align}
\nonumber
\beta^{p_k} = \mathrm{sigmoid}({w^r}^\top M^{p_k}_1).
\end{align}

\subsection{Answer Possibility Classifier}
\label{sec:classifier}

The classifier maps the output of the modeling layer 
to a probability for the answer possibility. It also takes the output for the first word, $M^{p_k}_1$, for all passages and concatenates them. 
Given $w^c \in \mathbb{R}^{Kd}$ as learnable parameters,
it calculates the answer possibility for the question as
\begin{align}
\nonumber
P(a) = \mathrm{sigmoid}({w^c}^\top [M^{p_1}_1; \ldots; M^{p_K}_1]).
\end{align}

\subsection{Answer Sentence Decoder}
\label{sec:decoder}

Given the outputs 
provided by the reader module, 
the decoder generates a sequence of answer words one element at a time. It is auto-regressive~\citep{Graves13}, consuming the previously generated words as additional input at each decoding step.

\subsubsection{Word Embedding Layer}

Let $y$ represent one-hot vectors of the words in the answer. This layer has the same components as the word embedding layer of the reader module, except that it uses a unidirectional ELMo to ensure that the predictions for position $t$ depend only on the known outputs at positions previous to $t$.

\paragraph{Artificial tokens.} 
To be able to use multiple answer styles within a single system, our model introduces an artificial token corresponding to the style at the beginning of the answer ($y_1$), as done in~\citep{JohnsonSLKWCTVW17,TakenoNY17}.
At test time, the user can specify the first token to control the style. This modification does not require any changes to the model architecture.  Note that introducing the token at the decoder prevents the reader module from depending on the answer style.

\subsubsection{Attentional Decoder Layer}
\label{sec:style}

This layer uses a stack of Transformer decoder blocks on top of the embeddings provided by the word embedding layer. The input is immediately mapped to a $d$-dimensional vector by a linear transformation, and the output is a sequence of $d$-dimensional vectors: $\{s_1, \ldots, s_T\}$.

\paragraph{Transformer decoder block.} 
In addition to the encoder block, this block consists of the second and third sub-layers after the self-attention block and before the feed-forward network, as shown in Figure~\ref{fig:model}. As in \citep{VaswaniSPUJGKP17}, the self-attention sub-layer uses a sub-sequent mask to prevent positions from attending to subsequent positions. The second and third sub-layers perform the multi-head attention over $M^q$ and $M^{p_\mathrm{all}}$, respectively. 
The $M^{p_\mathrm{all}}$ is the concatenated outputs of the encoder stack for the passages, 
\begin{align}
\nonumber
M^{p_\mathrm{all}} = [M^{p_1}, \ldots, M^{p_K}] \in \mathbb{R}^{d \times KL}.
\end{align}
Here, the $[,]$ operator denotes vector concatenation across the columns. 
This attention for the concatenated passages produces attention weights that are comparable between passages.

\subsubsection{Multi-source Pointer-Generator}
\label{sec:copy}

\begin{figure}[t!]
\centering
\includegraphics[width=.47\textwidth]{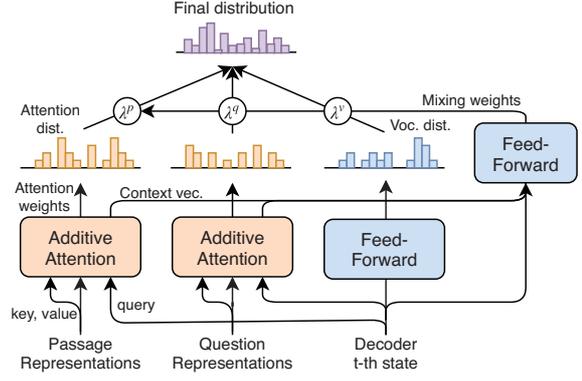}
\caption{Multi-source pointer-generator mechanism. 
For each decoding step $t$, mixture weights $\lambda^v, \lambda^q, \lambda^p$ 
for the probability of generating words from the vocabulary and copying words from the question and the passages are calculated.
The three distributions are weighted and summed to obtain the final distribution.
}
\label{fig:copy}
\end{figure}

Our extended mechanism allows both words to be generated from a vocabulary and words to be copied from both the question and multiple passages (Figure~\ref{fig:copy}). 
We expect that the capability of copying words will be shared among answer styles. 

\paragraph{Extended vocabulary distribution.}
Let the extended vocabulary,  $V_\mathrm{ext}$, be the union of the common words (a small subset of the full vocabulary, $V$, defined by the 
input-side word embedding matrix) and all words appearing in the input question and passages. $P^v$ then denotes the probability distribution of the $t$-th answer word, $y_t$, over the extended vocabulary. It is defined as:
\begin{align}
\nonumber
P^v(y_t)  =\mathrm{softmax}({W^2}^\top (W^1 s_t  + b^1)),
\end{align}
where the output embedding $W^2 \in \mathbb{R}^{d_\mathrm{word} \times V_\mathrm{ext}}$ is tied with the corresponding part of the input embedding~\citep{InanKS17}, and $W^1 \in \mathbb{R}^{d_\mathrm{word} \times d}$ and $b^1 \in \mathbb{R}^{d_\mathrm{word}}$ are learnable parameters. $P^v(y_t)$ is zero if $y_t$ is an out-of-vocabulary word for $V$.

\paragraph{Copy distributions.}
A recent Transformer-based pointer-generator
randomly chooses one of the attention-heads to form a copy distribution; 
that approach gave no significant improvements in text summarization~\citep{GehrmannDR18}.

In contrast, our model uses an additional attention layer for each copy distribution on top of the decoder stack. For the passages, the layer takes $s_t$ as the query and outputs 
$\alpha^p_t \in \mathbb{R}^{KL}$ as the attention weights and 
$c^p_t \in \mathbb{R}^d$ as the context vectors:
\begin{align}
\nonumber
e^{p_k}_l &= {w^p}^\top \tanh(W^{pm} M_l^{p_k} + W^{ps} s_t +b^p), \\
\label{eq:wordattn}
\alpha^p_t &= \mathrm{softmax}([e^{p_1}; \ldots; e^{p_K}]), \\
\nonumber
c^p_t &=  \textstyle \sum_{l} \alpha^p_{tl} M^{p_\mathrm{all}}_{l}, 
\end{align}
where $w^p, b^p \in \mathbb{R}^d$ and $W^{pm}, W^{ps} \in \mathbb{R}^{d \times d}$
are learnable parameters. 
For the question, our model uses another identical layer
and obtains $\alpha^q_t \in \mathbb{R}^J$ and $c^q_t \in \mathbb{R}^d$. 
As a result, $P^q$ and $P^p$ are the copy distributions over the extended vocabulary: \begin{align}
\nonumber
P^q(y_t) &=  \textstyle \sum_{j: x^q_j = y_t} \alpha^q_{tj},\\ \nonumber
P^p(y_t) &= \textstyle \sum_{l: x^{p_{k(l)}}_{l} = y_t} \alpha^p_{tl},
\end{align}
where $k(l)$ means the passage index corresponding to the $l$-th word in the concatenated passages.

\paragraph{Final distribution.}

The final distribution of $y_t$ is defined as a mixture of the three distributions:
\begin{align}
\nonumber
P(y_t) &= \lambda^v P^v(y_t) +  \lambda^q P^q(y_t) + \lambda^p P^p(y_t), \\
\nonumber
\lambda^v, \lambda^q, \lambda^p &= \mathrm{softmax}(W^m [s_t; c^q_t; c^p_t] + b^m),
\end{align}
where $W^m \in \mathbb{R}^{3 \times 3d}$ and $b^m \in \mathbb{R}^3$ are learnable parameters.

\subsubsection{Combined Attention}
\label{sec:combined}

In order not to attend words in irrelevant passages, our model introduces 
a combined attention. 
While the original technique combined 
word and sentence level attentions~\citep{HsuLLMTS18}, our model combines 
the word and passage level attentions. 
The word attention, Eq.~\ref{eq:wordattn}, is re-defined as
\begin{align}
\nonumber
\alpha^p_{tl} = \frac{\alpha^p_{tl} \beta^{p_{k(l)} }}{\sum_{l'} \alpha^p_{tl'} \beta^{p_{k(l')}}}.
\end{align}

\subsection{Loss Function}

We define the training loss as the sum of losses via 
\begin{align}
\nonumber
L(\theta) = L_\mathrm{dec} + \gamma_\mathrm{rank} L_\mathrm{rank} + \gamma_\mathrm{cls} L_\mathrm{cls}
\end{align}
where $\theta$ is the set of all learnable parameters, and $\gamma_\mathrm{rank}$ and $\gamma_\mathrm{cls}$ are balancing parameters.

The loss of the decoder, $L_\mathrm{dec}$, is the negative log likelihood of the whole target answer sentence averaged over $N_\mathrm{able}$ answerable examples:
\begin{align}
\nonumber
L_\mathrm{dec} = - \frac{1}{N_\mathrm{able}}\sum_{(a,y)\in \mathcal{D}} \frac{a}{T} \sum_t \log P(y_{t}),
\end{align}
where $\mathcal{D}$ is the training dataset.
The losses of the passage ranker, $L_\mathrm{rank}$, and
the answer possibility classifier, $L_\mathrm{cls}$, are 
the binary cross entropy between the true and predicted values averaged over all $N$ examples:
\begingroup\makeatletter\def\f@size{9.5}\check@mathfonts
\begin{gather}
\nonumber
L_\mathrm{rank} = -  \frac{1}{NK} \sum_k \sum_{r^{p_k}\in\mathcal{D}}  
\biggl(
\begin{split}
&r^{p_k} \log \beta^{p_k} +  \\
&(1-r^{p_k}) \log (1-\beta^{p_k}) 
\end{split}
\biggr),\\
\nonumber
L_\mathrm{cls} = - \frac{1}{N} \sum_{a \in \mathcal{D}} 
\biggl(
\begin{split}
&a \log P(a) + \\
&(1-a) \log (1-P(a)) 
\end{split}
\biggr).
\end{gather}
\endgroup

\section{Experiments on MS MARCO 2.1}

We evaluated our model on MS MARCO 2.1~\cite{Bajaj18}. 
It is the sole dataset providing abstractive answers with multiple styles and serves as a great test bed for building open-domain QA agents with the NLG capability that can be used in smart devices.
The details of our setup and output examples are in the supplementary material.

\subsection{Setup} 

\paragraph{Datasets.}
MS MARCO 2.1 provides two tasks for generative open-domain QA:
the \textbf{Q\&A} task and the Q\&A + Natural Language Generation (\textbf{NLG}) task.
Both tasks consist of questions submitted to Bing by real users, and each question refers to ten passages.
The dataset also includes annotations on the relevant passages, which were selected by humans to form the final answers, and on whether there was no answer in the passages.

\paragraph{Answer styles.}
We associated the two tasks with two answer styles.
The NLG task requires a well-formed answer that is an abstractive summary of the question and passages, averaging 16.6 words.
The Q\&A task also requires an abstractive answer but prefers 
it to be
more concise 
than in the NLG task, averaging 13.1 words, and many of the answers do not contain the context of the question.
For the question ``tablespoon in cup'', 
a reference answer in the Q\&A task is ``16,'' while 
that in the NLG task is ``There are 16 tablespoons in a cup.''

\begin{table}[t!]
\centering
{\small \tabcolsep=4.5pt
\begin{tabular}{c|c|ccc}
\hline
Dataset & Subset   & Train & Dev. & Eval. \\ \hline
             & ALL & 808,731 & 101,093 & 101,092\\
MS MARCO & ANS & 503,370 & 55,636 & --\\
             & NLG & 153,725 & 12,467 & --\\ \hline
NarrativeQA & Summary & 32,747 & 3,461 & 10,557 \\ \hline             
\end{tabular} \\
}
\caption{Numbers of questions used in the experiments.}
\label{tb:data}
\end{table}

\paragraph{Subsets.} In addition to the \textbf{ALL} dataset, we prepared two subsets for ablation tests as listed in Table~\ref{tb:data}. The \textbf{ANS} set consisted of answerable questions, and the \textbf{NLG} set consisted of the answerable questions and well-formed answers, so that NLG $\subset$ ANS $\subset$ ALL.
We note that multi-style learning enables our model to learn from different answer styles of data (i.e., the ANS set), and  multi-task learning with the answer possibility classifier enables our model to learn from both answerable and unanswerable data (i.e., the ALL set).

\paragraph{Training and Inference.}
We trained our model with mini-batches consisting of multi-style answers that were randomly sampled. 
We used a greedy decoding algorithm and did not use any beam search or random sampling, because they did not provide any improvements.

\paragraph{Evaluation metrics and baselines.}
ROUGE-L 
and BLEU-1 
were
used to evaluate the models' RC performance, where ROUGE-L is the main metric on the official leaderboard.
We used the reported scores of extractive~\citep{SeoKFH17,YanAAAI19,WuWLHWLLL18},  generative~\citep{TanWYDLZ18}, and unpublished RC models at the submission time.

In addition, to evaluate 
the individual contributions of our modules,
we used MAP and MRR
for the ranker and $F_1$ for the classifier, where the positive class was the answerable questions.

\subsection{Results} 

\paragraph{Does our model achieve state-of-the-art on the two tasks with different styles?}

\begin{table}
\centering
{\small \tabcolsep=4pt
\begin{tabular}{l|cc|cc}
\hline
 & \multicolumn{2}{c|}{NLG} & \multicolumn{2}{c}{Q\&A}\\ 
Model & R-L & B-1 & R-L & B-1\\ \hline
BiDAF$^a$ & 16.91 & 9.30 & 23.96 & 10.64 \\
Deep Cascade QA$^b$ & 35.14 & 37.35 &  52.01 & {\bf 54.64}\\ 
S-Net+CES2S$^c$ & 45.04 & 40.62 & 44.96 & 46.36\\
BERT+Multi-PGNet$^d$ & 47.37 & 45.09 & 48.14 & 52.03  \\
Selector+CCG$^e$ & 47.39 & 45.26 & 50.63 &52.03 \\ 
VNET$^f$ & 48.37 & 46.75 & 51.63 & 54.37\\ \hline
Masque (NLG; single) & 49.19 &  49.63 & 48.42 & 48.68\\
Masque (NLG; ensemble) & {\bf 49.61} & {\bf 50.13} & 48.92& 48.75 \\
\hline
Masque (Q\&A; single) & 25.66 & 36.62 & 50.93 & 42.37\\ 
Masque (Q\&A; ensemble) & 28.53 & 39.87 & {\bf 52.20} & 43.77 \\ 
\hline
Human Performance & 63.21 & 53.03  & 53.87 & 48.50\\
\hline
\end{tabular}
}
\caption{Performance of our and competing models on the MS MARCO V2 leaderboard 
(4 March 2019). 
$^a$\citet{SeoKFH17}; $^b$\citet{YanAAAI19};
$^c$Shao (unpublished), a variant of \citet{TanWYDLZ18};
$^d$Li (unpublished), a model using \citet{DevlinCLT18} and \citet{SeeLM17};
$^e$Qian (unpublished); 
$^f$\citet{WuWLHWLLL18}.
Whether the competing models are ensemble models or not is unreported. }

\label{tb:nlg-leaderboard}
\end{table}

Table~\ref{tb:nlg-leaderboard} shows the performance of our model and competing models on the leaderboard.
Our ensemble model of six training runs, where each model was trained with the two answer styles,
achieved state-of-the-art performance on both tasks in terms of ROUGE-L. 
In particular, for the NLG task, our single model outperformed competing models in terms of both ROUGE-L and BLEU-1.

\paragraph{Does multi-style learning improve the NLG performance?}

Table~\ref{tb:ablation} lists the results of an ablation test for our single model (controlled with the NLG style) on the NLG dev.~set\footnote{We confirmed with the organizer that the dev.~results
were much better than the test results, 
but there was no problem.}.
Our model trained with both styles outperformed the model trained with the single NLG style.
Multi-style learning enabled our model to improve its NLG
performance by also using non-sentence answers.

\begin{table}[t!]
\centering
{\small \tabcolsep=3pt 
\begin{tabular}{p{15em}|c|cc}
\hline
Model & Train & R-L & B-1 \\ \hline
Masque (NLG style; single)
& ALL&  {\bf 69.77} & {\bf 65.56} \\ 
\begin{tabular}{p{15em}}
w/o multi-style learning (\S\ref{sec:style})
\end{tabular} & NLG &68.20 & 63.95 \\
\begin{tabular}{p{15em}}
$\hookrightarrow$
w/o Transformer (\S\ref{sec:tfenc}, \S\ref{sec:style})
\end{tabular} & NLG & 67.13 & 62.96 \\ 
\begin{tabular}{p{15em}}
w/o passage ranker (\S\ref{sec:ranker})
\end{tabular} & NLG & 68.05 & 63.82 \\
\begin{tabular}{p{15em}}
w/o possibility classifier (\S\ref{sec:classifier})
\end{tabular} & ANS & 69.64 & 65.41 \\ \hline
Masque w/ gold passage ranker & ALL & 78.70 & 78.14 \\ 
\hline
\end{tabular}
}
\caption{Ablation test results 
on the NLG dev.~set. 
The models were trained with the subset listed in ``Train''.}
\label{tb:ablation}
\end{table}

\paragraph{Does the Transformer-based pointer-generator improve the NLG performance?}

Table~\ref{tb:ablation} shows that our model also outperformed 
the model that used RNNs and self-attentions instead of Transformer blocks as in MCAN~\citep{McCannKXS18}.
Our deep decoder captured the multi-hop interaction 
among the question, the passages, and the answer 
better than a single-layer LSTM decoder could.

\paragraph{Does joint learning with the ranker and classifier improve NLG performance?}

Furthermore, Table~\ref{tb:ablation} shows that 
our model (jointly trained with the passage ranker and answer possibility classifier) outperformed the model
that did not use the ranker and classifier.
Joint learning thus had a regularization effect on the question-passages reader.

We also confirmed that the gold passage ranker, which can perfectly predict the relevance of passages, significantly improved the RC performance. Passage ranking will be a key to developing a system that can outperform humans.

\paragraph{Does joint learning improve the passage ranking performance?}

\begin{table}[t!]
\centering
{\small \tabcolsep=3.5pt
\begin{tabular}{p{14em}|c|cc}
\hline
Model & Train & MAP & MRR \\ \hline
Bing (initial ranking) & - & 34.62 & 35.00 \\ \hline
Masque (single) &ALL & {\bf 69.51} & {\bf 69.96}\\ 
\begin{tabular}{p{14em}}
w/o answer decoder (\S\ref{sec:decoder})
\end{tabular} & ALL & 67.03 & 67.49 \\ 
\begin{tabular}{p{14em}}
w/o multi-style learning (\S\ref{sec:style})
\end{tabular} & NLG & 65.51 & 65.59 \\ 
\begin{tabular}{p{14em}}
w/o possibility classifier (\S\ref{sec:classifier})
\end{tabular} & ANS & 69.08 & 69.54 \\ 
\hline
\end{tabular}
}
\caption{Passage ranking results on the ANS dev.~set. }
\label{tb:ranker}
\end{table}

Table~\ref{tb:ranker} lists the passage ranking performance 
on the ANS dev.~set\footnote{This evaluation requires our ranker to re-rank 10 passages. It is not the same as the Passage Re-ranking task.}.
The ranker shares the 
question-passages 
reader with the answer decoder, and this sharing contributed to improvements over the ranker trained without the answer decoder. 
Also, our ranker outperformed the initial ranking provided by Bing by a significant margin.

\begin{figure}[t!]
\centering
\includegraphics[width=.36\textwidth]{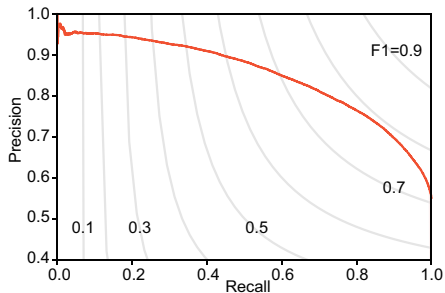}
\caption{Precision-recall curve for answer possibility classification on the ALL dev.~set.}
\label{fig:anspos}
\end{figure}

\paragraph{Does our model accurately identify answerable questions?}

Figure~\ref{fig:anspos} shows the precision-recall curve for answer possibility classification on the ALL dev.~set.
Our model identified the answerable questions well. The maximum 
$F_1$ score was 0.7893,  where the threshold of answer possibility was $0.4411$.
This is the first report on answer possibility classification with MS MARCO 2.1.

\begin{figure}[t!]
\centering
\includegraphics[width=.47\textwidth]{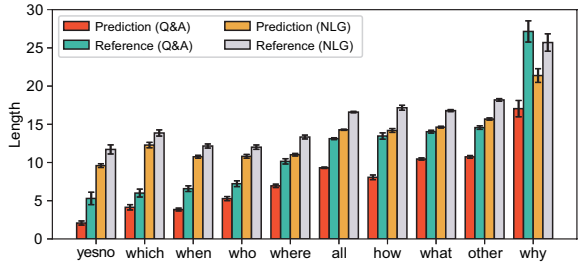}
\caption{Lengths of answers generated by Masque broken down by the answer style and query type on the NLG dev.\ set. The error bars indicate standard errors.}
\label{fig:length}
\end{figure}

\paragraph{Does our model control answer lengths with different styles?}
Figure~\ref{fig:length} shows the lengths of the answers generated by our model broken down by the answer style and query type.
The generated answers were relatively shorter than the reference answers, especially for the Q\&A task, but well controlled with the target style for every query type.  
The short answers degraded our model's BLEU scores in the 
Q\&A task (Table~\ref{tb:nlg-leaderboard}) 
because of BLEU's brevity penalty~\cite{PapineniRWZ02}.

\section{Experiments on NarrativeQA}

Next, we evaluated our model on NarrativeQA~\cite{KociskySBDHMG18}. 
It requires understanding the underlying narrative rather than relying on shallow pattern matching. 
Our detailed setup and output examples are in the supplementary material.

\subsection{Setup}
We only describe the settings specific to this experiment.

\paragraph{Datasets.}
Following previous studies, we used the summary setting for the comparisons with the reported baselines, where each question refers to one summary (averaging 659 words), and there is no unanswerable questions.
Our model therefore did not use the passage ranker and answer possibility classifier. 

\paragraph{Answer styles.}
The NarrativeQA dataset does not explicitly provide multiple answer styles.
In order to evaluate the effectiveness of multi-style learning, we used the NLG subset of MS MARCO as additional training data.
We associated the NarrativeQA and 
NLG datasets with two answer styles. 
The answer style of NarrativeQA (\textbf{NQA}) is different from that of MS MARCO (\textbf{NLG}) in that the answers are short (averaging 4.73 words) and contained frequently pronouns.
For instance, for the question ``Who is Mark Hunter?'', 
a reference is ``He is a high school student in Phoenix.''

\paragraph{Evaluation metrics and baselines.}
BLEU-1 and 4,
METEOR, 
and ROUGE-L were used in accordance with the evaluation in the dataset paper~\citep{KociskySBDHMG18}.
We used the reports of top-performing 
extractive~\citep{SeoKFH17,
TayLHS18,HuPWHLYZ18} and generative~\citep{BauerWB18,IndurthiYBC18}  
models.

\subsection{Results} 

\begin{table}
\centering
{\small \tabcolsep=4.5pt
\begin{tabular}{l|cccc}
\hline
Model & B-1 & B-4 & M & R-L \\ \hline
BiDAF$^a$ & 33.72 & 15.53 & 15.38 & 36.30 \\ 
DECAPROP$^b$ & 42.00 & 23.42 & 23.42 & 40.07 \\
MHPGM+NOIC$^c$ & 43.63 & 21.07 & 19.03 & 44.16 \\
ConZNet$^d$ & 42.76 & 22.49 & 19.24 & 46.67 \\
RMR+A2D$^e$ & 50.4 & 26.5 & N/A & 53.3 
\\ \hline
Masque (NQA) & {\bf 54.11} &{\bf 30.43} & {\bf 26.13} & {\bf 59.87}\\ 
\hspace{0.5em} w/o multi-style learning & 48.70 & 20.98 & 21.95 & 54.74 \\ 
Masque (NLG) & 39.14 & 18.11 & 24.62 & 50.09 \\ 
\hline
\hline
Masque (NQA; valid.)$^f$ & 52.78 & 28.72 & 25.38 & 58.94\\\hline
\end{tabular}
}
\caption{Performance of our and competing models on the NarrativeQA test set. 
$^a$\citet{SeoKFH17}; 
$^b$\citet{TayLHS18};
$^c$\citet{BauerWB18};
$^d$\citet{IndurthiYBC18};
$^e$\citet{HuPWHLYZ18}.
$^f$Results on the NarrativeQA validation set.
}
\label{tb:narrative-test-leaderboard}
\end{table}

\paragraph{Does our model achieve state-of-the-art performance?} 
Table~\ref{tb:narrative-test-leaderboard} shows that 
our single model, trained with two styles
and controlled with the NQA style, 
pushed forward the
state-of-the-art by a significant margin. 
The evaluation scores of the model controlled with the NLG style were low because the two styles are different. 
Also, our model without multi-style learning (trained with only the NQA style) outperformed the baselines
in terms of ROUGE-L. 
This indicates that our model architecture itself is powerful for natural language understanding in RC.

\section{Related Work and Discussion}

\paragraph{Transfer and multi-task learning in RC.}

Recent breakthroughs in transfer learning demonstrate that 
pre-trained language models 
perform well on RC
with minimal modifications~\citep{PetersNIGCLZ18,DevlinCLT18,RadfordNSS18,Radford19}. 
In addition, our model also uses ELMo~\citep{PetersNIGCLZ18} for contextualized embeddings.

Multi-task learning is a transfer mechanism to improve generalization performance~\citep{Caruana97}, and it is generally applied by sharing the hidden layers between all tasks, while keeping task-specific layers.
\citet{WangAAAI2018} and \citet{NishidaSOAT18} reported that 
the sharing of the hidden layers between the multi-passage RC and passage ranking tasks was effective.
Our results also showed the effectiveness of 
the sharing of the question-passages reader 
module among the RC, passage ranking, and answer possibility classification tasks.

In multi-task learning without task-specific layers,
\citet{DevlinCLT18} and \citet{ChenFWB17} improved RC performance by learning multiple datasets from the same extractive RC
setting. 
\citet{McCannKXS18} and \citet{Yogatama19} investigated multi-task and curriculum learning 
on many different NLP tasks; their results were below task-specific RC models.
Our multi-style learning does not use style-specific layers; instead uses a style-conditional decoder. 

\paragraph{Generative RC.} 

S-Net \citep{TanWYDLZ18} 
used an extraction-then-synthesis mechanism for multi-passage RC. 
The models proposed by \citet{McCannKXS18}, \citet{BauerWB18}, and \citet{IndurthiYBC18} used an RNN-based pointer-generator mechanism for single-passage RC.
Although these mechanisms can alleviate the lack of training data, 
large amounts of data are still required. Our multi-style learning will be a key technique enabling learning from many RC datasets with different styles.

In addition to MS MARCO and NarrativeQA, 
there are other datasets that provide abstractive answers. 
DuReader~\citep{HeLLLZXLWWSLWW18}, a Chinese multi-document RC
dataset, provides longer documents and answers than those of MS MARCO.
DuoRC~\citep{KhapraSSA18} and CoQA~\citep{ReddyCM18} contain abstractive answers;  most of the answers are 
short phrases.

\paragraph{Controllable text generation.}

Many studies have been carried out in the framework of style transfer, which is the task of rephrasing a text so that it contains specific styles such as sentiment. 
Recent studies have used artificial tokens~\citep{SennrichHB16,JohnsonSLKWCTVW17}, variational auto-encoders~\citep{HuYLSX17}, or adversarial training~\citep{FuTPZY18,TsvetkovBSP18}
to separate the content and style on the encoder side. 
On the decoder side, conditional language modeling has been used to generate output sentences with the target style.
In addition,
output length control with conditional language modeling has been well studied~\citep{KikuchiNSTO16,TakenoNY17,FanGA18}. Our style-controllable RC relies on conditional language modeling in the decoder. 

\paragraph{Multi-passage RC.} 

The simplest approach is to concatenate the passages and find the answer from the concatenation, as in \citep{WangYWCZ17}. Earlier pipelined models found a small number of relevant passages with a TF-IDF based ranker and passed them to a neural reader~\citep{ChenFWB17,GardnerC18}, while more recent 
models have used a neural re-ranker to more accurately select the relevant passages~\citep{WangAAAI2018,NishidaSOAT18}. 
Also, non-pipelined models (including ours) consider all the provided passages and find the answer by comparing scores between passages~\citep{TanWYDLZ18,WuWLHWLLL18}.  The most recent models make a proper trade-off between efficiency and accuracy~\citep{YanAAAI19,MinZSX18}.

\paragraph{RC with unanswerable question identification.}

The previous work of \citep{LevySCZ17,GardnerC18} outputted a no-answer score depending on the probability of all answer spans. \citet{HuWPHYZ18}~proposed an answer verifier to compare an answer with the question. \citet{SunLQL18} 
jointly learned an RC model and an answer verifier. Our model introduces a classifier on top of the question-passages reader, which is not dependent on the generated answer. 
\paragraph{Abstractive summarization.}

Current state-of-the-art models use the pointer-generator mechanism~\citep{SeeLM17}.
In particular, content selection approaches, which decide what to summarize, have recently been used with abstractive models. Most methods select content at the sentence level~\citep{HsuLLMTS18,ChenB18} or the word level \citep{PasunuruB18,LiXLG18,GehrmannDR18}. Our model incorporates content selection at the passage level in the combined attention. 

Query-based 
summarization has rarely been studied because of a lack of datasets. \citet{NemaKLR17} proposed an attentional encoder-decoder model; however, \citet{KhapraSSA18} reported that it performed worse than BiDAF on DuoRC.
\citet{HasselqvistHK17} proposed a pointer-generator based model; however, it does not consider copying words from the question. 

\section{Conclusion}

This study sheds light on multi-style generative RC.
Our proposed model, \textit{Masque}, is based on multi-source abstractive summarization
and learns multi-style answers together.
It achieved state-of-the-art performance on 
the Q\&A task and the Q\&A + NLG task of
MS MARCO 2.1 and the summary task of NarrativeQA.
The key to its success is transferring the style-independent NLG capability to the target style 
by use of
the question-passages reader and the conditional pointer-generator decoder.
In particular, the capability of copying words from the question and passages can be shared among the styles, 
while the capability of controlling the mixture weights for the generative and copy distributions can be acquired for each style. 
Our future work will involve exploring the potential of our multi-style learning towards natural language understanding.

\bibliographystyle{acl_natbib}
\bibliography{references}



\appendix

\section{Supplementary Material}

\subsection{Experimental Setup for MS MARCO}
\label{sec:setup1}

\paragraph{Model configurations.}
We trained our model on a machine with eight NVIDIA P100 GPUs. Our best model was jointly trained with the two answer styles in the ALL set for a total of eight epochs with a batch size of 80, where each batch consisted of multi-style answers that were randomly sampled.
The training took roughly six days.  
The hidden size $d$ was 304, and the number of attention heads was 8. The inner state size of the feed-forward networks was 256. The numbers of shared encoding blocks, modeling blocks for a question, modeling blocks for passages, and decoder blocks were 3, 2, 5, and 8, respectively. We used the pre-trained uncased 300-dimensional 
GloVe~\citep{PenningtonSM14}%
\footnote{\burl{https://nlp.stanford.edu/projects/glove/}}
and the original 512-dimensional 
ELMo~\citep{PetersNIGCLZ18}%
\footnote{\burl{https://allennlp.org/elmo}}. 
We used the spaCy tokenizer, and all input words were lowercased except the input for ELMo. The output words were also lowercase.
The number of common words 
in $V_\mathrm{ext}$ in the extended vocabulary 
was 5,000. Each passage and each answer were truncated to 100 words for training.

\paragraph{Optimizer.}
We used Adam~\citep{KingmaB15} with $\beta_1 = 0.9$, $\beta_2 = 0.999$, and $\epsilon = 10^{-8}$. The weights were initialized using $N(0, 0.02)$, except that the biases of all the linear transformations were initialized with zero vectors. The learning rate was increased linearly from zero to $2.5 \times 10^{-4}$ in the first 2,000 steps and then annealed to 0 by using a cosine schedule. All parameter gradients were clipped to a maximum norm of $1$. An exponential moving average was applied to all trainable variables with a decay rate of 0.9995. The balancing factors for joint learning, $\lambda_\mathrm{rank}$ and $\lambda_\mathrm{cls}$, were set to 0.5 and 0.1, respectively.

\paragraph{Regularization.}

We used a modified version of the L$_2$ regularization proposed in~\citep{LoshchilovH17}, with $w = 0.01$ on all non-bias. 
We additionally used a dropout~\citep{SrivastavaHKSS14} rate of 0.3 for all highway networks and residual and scaled dot-product attention operations in the multi-head attention mechanism. We also used one-sided label smoothing~\citep{SzegedyVISW16} for the passage relevance and answer possibility labels. We smoothed only the positive labels to 0.9.


\paragraph{Ensemble model.}
The ensemble model consisted of six training runs with identical architectures and hyperparameters but with different weight initializations.
The final answer was decided with a weighted majority, where we used the ROUGE-L score for the dev.~set as the weight of each model.


\paragraph{Evaluation settings.}
We used the official evaluation script. 
The answers were normalized by making words lowercase. 

\subsection{Experimental Setup for NarrativeQA}
\label{sec:setup2}

\paragraph{Model configurations.}
Our best model was jointly trained with 
 the NarrativeQA and MS MARCO NLG datasets for a total of seven epochs with a batch size of 64, where each batch 
 consisted of multi-style answers that were randomly sampled.
 For efficient multi-style learning, each summary in the NarrativeQA dataset was divided into ten passages (size of 130 words) with sentence-level overlaps such that each sentence in the summary was entirely contained in a passage.  Each passage from MS MARCO was also truncated to 130 words.
The rest of the configuration was the same as in the MS MARCO experiments.

\paragraph{Evaluation settings.}

An official evaluation script is not provided, so we used the evaluation script created by \citet{BauerWB18}%
\footnote{\burl{https://github.com/yicheng-w/CommonSenseMultiHopQA/}}. 
The answers were normalized 
by
making words lowercase and removing punctuation marks.

\subsection{Output Examples Generated by Masque}

Tables \ref{tb:examples} and \ref{tb:examples2} list the 
generated examples for questions from MS MARCO 2.1 and NarrativeQA, respectively.  We can see from the examples that our model could control answer styles appropriately for various question and reasoning types.
We did find some important errors: 
style errors, yes/no classification errors, copy errors with respect to numerical values, grammatical errors, and multi-hop reasoning errors. 

\onecolumn
\label{sec:examples}

\begin{table*}[h!]
\centering
{\footnotesize
\tabcolsep=1pt
\vspace{0.5pt}
\begin{tabular}{p{50em}}
\hline
\vspace{0.5pt}
\pbox{50em}{
\textbf{(a) Question}: why your body would feel like it is shaking\\
\textbf{Relevant Passage}: Shaking is a symptom in which a person has tremors (shakiness or small back and forth movements) in part or all of his body. \textcolor{black}{Shaking can be due to cold body temperatures, rising fever (such as with infections), neurological problems, medicine effects, drug abuse, etc.} ...Read more. \\
\textbf{Reference Answer (Q\&A)}: Shaking can be due to cold body temperatures, rising fever (such as with infections), neurological problems, medicine effects, drug abuse, etc.  \\
\textbf{Prediction (Q\&A)}: because of cold body temperatures , rising fever , neurological problems , medicine effects , drug abuse~.~\cmark\\
\textbf{Reference Answers (NLG)}: Shaking can be due to cold body temperatures, rising fever, neurological problems, medicine effects and drug abuse. / 
Body would feel like it is shaking due to cold body temperatures, rising fever, neurological problems, medicine effects, drug abuse. \\\
\textbf{Prediction (NLG)}: your body would feel like it is shaking because of cold body temperatures , rising fever , neurological problems , medicine effects , drug abuse . \cmark
}
\vspace{1pt}
\\ \hline
\vspace{0.5pt}
\pbox{50em}{
\textbf{(b) Question}: \_\_\_\_\_ is the name used to refer to the era of legalized segregation in the united states \\
\textbf{Relevant Passage}: \textcolor{black}{Jim Crow law, in U.S. history, any of the laws that enforced racial segregation in the South between the end of Reconstruction in 1877 and the beginning of the civil rights movement in the 1950s.}
\\
\textbf{Reference Answer (Q\&A)}: Jim Crow \\
\textbf{Prediction (Q\&A)}: jim crow \cmark \\
\textbf{Reference Answer (NLG)}: Jim Crow is the name used to refer to the era of legalized segregation in the United States. \\
\textbf{Prediction (NLG)}: jim crow is the name used to refer to the era of legalized segregation in the united states . \cmark
}
\vspace{1pt}
\\ \hline
\vspace{0.5pt}
\pbox{50em}{
\textbf{(c) Question}: average height nba player\\
\textbf{Relevant Passage}: \textcolor{black}{The average height of an NBA player is around 6 feet 7 inches.} The tallest NBA player ever was Gheorghe Muresan, who was 7 feet 7 inches tall. In contrast, the shortest NBA player ever was Tyrone Muggsy Bogues, who was 5 feet 3 inches tall. \\
\textbf{Reference Answer (Q\&A)}: Around 6 feet 7 inches \\
\textbf{Prediction (Q\&A)}: 6 feet 7 inches	\cmark \\
\textbf{Reference Answers (NLG)}: The average height of NBA players is around 6 feet, 7 inches. / The height of NBA player is around 6 feet 7 inches.\\
\textbf{Prediction (NLG)}: the average height of \underline{an national} basketball association player is 6 feet 7 inches . \cmark
}
\vspace{1pt}
\\ \hline 
\vspace{0.5pt}
\pbox{50em}{
\textbf{(d) Question}: population of henryetta ok\\
\textbf{Relevant Passage}: \textcolor{black}{With it's population of 5,927 people, Henryetta represented 14.79\% of the county's total.} In addition, where the city of Henryetta has a population density of 891 people per square mile [344.2 people/km2], Okmulgee County averages 57.5 people per square mile [22.19 people/km2].\\
\textbf{Reference Answer (Q\&A)}: 5927 \\
\textbf{Prediction (Q\&A)}: the population of henryetta , oklahoma is 5,927 .   \cmark \hspace{.5em} (content is OK, but style is NG \xmark)\\
\textbf{Reference Answer (NLG)}: The population of Henryetta, Oklahoma is 5,927. \\
\textbf{Prediction (NLG)}:  the population of henryetta , oklahoma is 5,927 . \cmark
}
\vspace{1pt}
\\ \hline 
\vspace{0.5pt}
\pbox{50em}{
\textbf{(e) Question}: does gameplay programmer need math skill\\
\textbf{Relevant Passage}: \textcolor{black}{A good computer programmer is more of a problem solver and logical thinker than a math buff.} And besides, the industry is peppered with many successful computer programmers who do not really know much about mathematics. \\
\textbf{Reference Answer (Q\&A)}: No \\
\textbf{Prediction (Q\&A)}: yes \xmark \\
\textbf{Reference Answers (NLG)}: No, a gameplay programmer doesn't need math skill. / No, gameplay programmer do not need an math skill.\\
\textbf{Prediction (NLG)}: no , \underline{gameplay programmer does} not need math skill . \cmark
}
\vspace{1pt}
\\ \hline
\vspace{0.5pt}
\pbox{50em}{
\textbf{(f) Question}: how long does a freezer take to cool down\\
\textbf{Relevant Passage}: Quick Answer. \textcolor{black}{It takes anywhere from three to 24 hours for a refrigerator to reach safe temperatures for storing food, depending on the size and type of unit.} When the refrigerator compartment reaches 40 degrees Fahrenheit and the freezer reaches 5 degrees Fahrenheit, it is safe to transfer food items. Keep Learning. \\
\textbf{Reference Answer (Q\&A)}: 24 hours\\
\textbf{Prediction (Q\&A)}: 4 to 5 hours \xmark \\
\textbf{Reference Answers (NLG)}: A freezer takes 24 hours to cool down. / A  freezer take to cool down is 24 hours.\\
\textbf{Prediction (NLG)}: a freezer takes 4 to 12 hours to cool down . \xmark
}
\vspace{1pt}
\\ \hline
\end{tabular}
}
\caption{Output examples generated by Masque from MS MARCO.  The model was trained with the Q\&A and NLG styles. The relevant passage is one that an annotator selected to compose the reference answer. The model could control answer styles appropriately for (a) natural language, (b) cloze-style, and (c) keywords questions. (d) The answer style 
was incorrect. (e)
The answers were not consistent between the styles. (f) Copying from numerical words worked poorly. There were some grammatical errors in the generative answers, which are \underline{underlined}. 
}
\label{tb:examples}
\end{table*}

\onecolumn
\label{sec:examples2}

\begin{table*}[h!]
\centering
{\footnotesize
\tabcolsep=1pt
\vspace{0.5pt}
\begin{tabular}{p{50em}}
\hline
\vspace{0.5pt}
\pbox{50em}{ 
\textbf{(a) Question}:  Where does Mark broadcast his radio station? \\
\textbf{Summary}: \textcolor{black}{Mark Hunter (Slater), a high school student in a sleepy suburb of Phoenix, Arizona, starts an FM pirate radio station that broadcasts from the basement of his parents' house.} Mark is a loner, an outsider, whose only outlet for his teenage angst and aggression is his unauthorized radio station. 
His pirate station's theme song is "Everybody Knows" by Leonard Cohen and there are glimpses of cassettes by such alternative musicians as The Jesus and Mary Chain, Camper Van Beethoven, Primal Scream, Soundgarden, Ice-T, Bad Brains, Concrete Blonde, Henry Rollins, and The Pixies. By day, Mark is seen as a loner, hardly talking to anyone around him; by night, he expresses his outsider views about what is wrong with American society. When he speaks his mind about what is going on at his school and in the community, more and more of his fellow students tune in to hear his show. (...) \\
\textbf{Reference Answers}:  In his parent's basement. / His parents' basement. \\
\textbf{Prediction (NQA)}:  the basement of his parents ' house \cmark\\
\textbf{Prediction (NLG)}: \underline{mark broadcast} his radio station in the basement of his parents ' house . \cmark
}
\vspace{1pt}
\\ \hline
\vspace{0.5pt}
\pbox{50em}{ 
\textbf{(b) Question}: Fletch is a reporter for what newspaper? \\
\textbf{Summary}:  \textcolor{black}{Los Angeles Times reporter Irwin "Fletch" Fletcher (Chase) is writing an article exposing drug trafficking on the beaches of Los Angeles.} Posing as an addict during his investigation, he is approached by Boyd Aviation executive vice president Alan Stanwyk (Matheson) who mistakenly assumes Fletch is a junkie. 
Stanwyk claims to have bone cancer, with only months left to live, and wishes to avoid the pain and suffering. Stanwyk offers \$50,000 for Fletch to come to his mansion in a few days time, kill him, and then escape to Rio de Janeiro, staging the murder to look like a burglary. Fletch, while not completely convinced on the truth of Stanwyk's story, reluctantly agrees to the plan. Along with his colleague Larry (Davis), he begins investigating Stanwyk instead of completing his drug trafficking exposĂŠ, much to the disapproval of his overbearing editor Frank Walker (Libertini). Disguised as a doctor, Fletch accesses Stanwyk's file at the hospital and learns Stanwyk lied about having cancer. (...) \\
\textbf{Reference Answers}: Los Angeles Times / Los Angeles \\
\textbf{Prediction (NQA)}: los angeles times \cmark \\
\textbf{Prediction (NLG)}: fletch is a reporter for los angeles times .  \cmark
}
\vspace{1pt}
\\ \hline
\vspace{0.5pt}
\pbox{50em}{ 
\textbf{(c) Question}:  How long approximately was the voyage from London to Thailand supposed to take? \\
\textbf{Summary}: 
(...) The story is set twenty-two years earlier, when Marlow was 20. With two years of experience, most recently as third mate aboard a crack clipper, Marlow receives a billet as second mate on the barque Judea. The skipper is Captain John Beard, a man of about 60. 
 This is Beard's first command. The Judea is an old boat, belonging to a man "Wilmer, Wilcox or something similar", suffering from age and disuse in Shadewell basin. \textcolor{black}{The 400-ton ship is commissioned to take 600 tons of coal from England to Thailand. The trip should take approximately 150 days. The ship leaves London loaded with sand ballast and heads north to the Senn river to pick up the cargo of coal.} On her way, the Judea suffers from her ballast shifting aside and the crew go below to put things right again. The trip takes 16 days because of inclement weather, and the battered ship must use a tug boat to get into port. The Judea waits a month on the Tyne to be loaded with coal. 
The night before she ships out she is hit by a steamer, the Miranda or the Melissa. The damage takes another three weeks to repair. Three months after leaving London, the Judea ships off for Bangkok. The Judea travels through the North Sea and Britain. 300 miles west of the Lizard a winter storm, 'the famous winter gale of twenty-two years ago', hits. (...) \\
\textbf{Reference Answers}:  Approximately 150 days / 150 days \\
\textbf{Prediction (NQA)}: 150 days \cmark \\
\textbf{Prediction (NLG)}: the voyage from london to thailand was supposed to take 150 days . \cmark
}
\vspace{1pt}
\\ \hline
\vspace{0.5pt}
\pbox{50em}{ %
\textbf{(d) Question}: Why does Jamie start avoiding Landon? \\
\textbf{Summmary}: 
(...) During these functions, Landon notices Jamie Sullivan, a girl he has known since kindergarten and who has attended many of the same classes as him, and is also the local minister's daughter. Since he's one of the in-crowd, he has seldom paid any attention to Jamie, who wears modest dresses and owns only one sweater. Jamie is labeled an outsider and a geek. She makes no attempt to wear make-up or otherwise improve her looks or attract attention to herself. Landon has trouble learning his lines for the play. Jamie, who is also in the play, agrees to help him on one condition: Jamie warns Landon not to fall in love with her; he laughs it off and dismisses it as a foolish idea. Landon and Jamie begin practicing together at her house after school. They get to know each other and a spark of affection arises between them. On the opening night of the play, Jamie astounds Landon and the entire audience with her beauty and her voice. Onstage at the peak of the ending to the play, Jamie sings. \textcolor{black}{When Jamie finishes, Landon improvises and kisses her which is not a part of the play. Afterwards, Jamie avoids Landon, and it is not until Landon's friends play a cruel prank on Jamie and he protects her in opposition to his friends that she warms up to him again.} Landon asks Jamie on a date soon after, but Jamie says her father doesn't allow her to date. (...) \\
\textbf{Reference Answers}: Because he kissed her in the play. / He kisses her \\
\textbf{Prediction (NQA)}: he is not a part of the play \xmark \\
\textbf{Prediction (NLG)}: he is not a part of the play \xmark
}
\vspace{1pt}
\\ \hline
\end{tabular}
}
\caption{Output examples generated by Masque from NarrativeQA. The model was trained with the NarrativeQA (NQA) and MS MARCO (NLG) styles. It could control answer styles appropriately for questions that required (a,b) single-sentence reasoning
and (c) multi-sentence reasoning. (d) Example of an error in multi-sentence reasoning. There were some grammatical errors in the generative answers, which are \underline{underlined}. 
}
\label{tb:examples2}
\end{table*}

\end{document}